\documentclass[letterpaper, 10 pt, conference]{ieeeconf}  

\IEEEoverridecommandlockouts                              

\usepackage[utf8]{inputenc} 
\usepackage[T1]{fontenc}    
\usepackage{hyperref}       
\usepackage{url}            
\usepackage{booktabs}       
\usepackage{amsfonts}       
\usepackage{nicefrac}       
\usepackage{microtype}      

\usepackage{amsmath,amssymb,mathrsfs}
\usepackage[ruled, linesnumbered, vlined]{algorithm2e}
\usepackage{comment}
\usepackage{graphicx}
\usepackage{graphics}
\usepackage[tight]{subfigure}
\usepackage{wrapfig}
\usepackage{float}
\usepackage[table]{xcolor}
\usepackage{tabu}
\usepackage{multirow}
\usepackage{array}
\usepackage{enumerate}
\usepackage{sidecap}
\usepackage{flushend}
\usepackage{hyperref}
\usepackage{color,soul}
\usepackage{url}
\usepackage{changepage}  
\graphicspath{{./figs/}}

\let\labelindent\relax
\usepackage[shortlabels]{enumitem}
                    \setlist[enumerate, 1]{1\textsuperscript{o}}

\usepackage{color}

\definecolor{lightblue}{rgb}{0,0.2,1}
\definecolor{black}{rgb}{0,0,0}
\hypersetup{
    colorlinks=true,%
    citebordercolor=white,%
    filebordercolor=white,%
    linkbordercolor=white,%
    urlbordercolor=white,%
    citecolor=black,%
    linkcolor=black,%
    urlcolor=black%
}

\newcounter{tecounter}
\newenvironment{spmatrix}[1]
 {\def\mysubscript{#1}\mathop\bgroup\begin{pmatrix}}
 {\end{pmatrix}\egroup_{\textstyle\mathstrut\mysubscript}}

\title{\LARGE \bf
Accelerating Goal-Directed Reinforcement Learning by Model Characterization
}

\author{Shoubhik Debnath$^{1}$, Gaurav Sukhatme$^{2}$, Lantao Liu$^{3}$
\thanks{$^{1}$Shoubhik Debnath is with NVIDIA Corporation, Santa Clara, CA 95051, USA. E-mail:
        {\tt\small sdebnath@nvidia.com}}%
\thanks{$^{2}$Gaurav Sukhatme is with the Department of Computer Science at the University of Southern California, Los Angeles, CA 90089, USA. E-mail:
        {\tt\small gaurav@usc.edu}}%
\thanks{$^{3}$Lantao Liu is with the Intelligent Systems Engineering Department at  Indiana University - Bloomington,
        Bloomington, IN 47408, USA. E-mail:
        {\tt\small lantao@iu.edu}}%
\thanks{The paper was published in 2018 IEEE/RSJ International Conference on
Intelligent Robots and Systems (IROS).}
}

\begin{document}

\maketitle

\begin{abstract} 
We propose a hybrid approach aimed at improving the sample efficiency in goal-directed reinforcement learning. We do this via a two-step mechanism where firstly, we approximate a model from Model-Free reinforcement learning. Then, we leverage this approximate model along with a notion of reachability using Mean First Passage Times to perform Model-Based reinforcement learning.  Built on  such  a  novel  observation,  we  design  two  new  algorithms - Mean First Passage Time based Q-Learning (MFPT-Q) and Mean  First  Passage  Time  based DYNA (MFPT-DYNA), that have been fundamentally modified from the state-of-the-art reinforcement learning techniques. Preliminary results have shown that our hybrid approaches converge with much fewer iterations than their corresponding state-of-the-art counterparts and therefore requiring much fewer samples and much fewer training trials to converge. 
\end{abstract}



\section{Introduction}

Reinforcement Learning (RL) has been successfully applied to numerous challenging problems for autonomous agents to behave intelligently in unstructured real-world environment. One interesting area of research in RL which motivates this work is goal-directed reinforcement learning problem (GDRLP) \cite{Braga1998} \cite{Koenig1996}. In GDRLP, the learning process takes place in two stages. The first stage focuses on solving the goal-directed exploration problem (GDEP) which allows an agent to determine a viable path from an initial state to a goal state in an unknown or only partially known state space. The path found in this stage need not be the optimal one. In the second stage, the agent takes advantage of the previously learned knowledge to optimize the path to the goal state. 
The two stages iterate in order to converge to the action policy.

RL methods are generally divided into Model-Free (MF) and Model-Based (MB) approaches. MF methods can learn complex policies with minimal feature and policy engineering work. However, the convergence of such methods might require millions of trials and hence they are sample inefficient \cite{Kober2013} \cite{Schulman2015}. MB methods require much smaller number of real-world trials to converge but need an accurate model of real-world physical system and the environment which might be challenging to obtain \cite{Deisenroth2013}. 
Also, relying on an accurate model can be problematic because small glitches on the model may lead to catastrophic consequences.

In this paper, we leverage the benefits of both approaches - MB and MF, with an aim to improve the sample efficiency during the training process. We achieve this via a two-step iterative learning mechanism. In the first step,  we learn an approximate model of the physical system using an MF scheme. 
Note that, our approach does not need to construct an accurate model, instead a rough model with little cost is sufficient which is used to characterize the structure of the problem. 
In the second step, we leverage this approximate model along with the notion of reachability using Mean First Passage Times (MFPT) where the result is used to guide following MF exploration and learning.

\textbf{Contribution: } 
The main contributions of our work are:

\begin{itemize}
  \item We propose a hybrid RL approach that introduces a model-based characterization into the state-of-the-art RL algorithms to improve sample-efficiency.
  \item The model-based characterization is achieved via a model-based RL algorithm that is robust to an approximate model for learning complex policies.
\end{itemize}

We demonstrate our proposed method on two tasks related to the path-planning domain in 2D and 3D simulation environments respectively. The goal of the agent in both tasks is to learn an optimal policy to reach a goal state from a given start location. Our results show that the proposed hybrid algorithms with model-based characterization were able to learn the optimal policy in very few trials, thereby improving the sample efficiency and accelerating the learning process.

\vspace{-8pt}
\section{Related Work} 

Earlier works have demonstrated various approaches to solve goal-directed reinforcement learning problem. Braga et al. presented a solution to solve GDRLP for an indoor unknown environment. Firstly, using temporal difference learning method, they find an initial solution to reach the goal and then improve upon the initial solution by employing a variable learning rate \cite{Braga1998}. Several earlier works also focused on goal-directed exploration as it provides insight into the corresponding RL problems. It has been proved that goal-directed exploration with RL methods can be intractable, therefore demonstrating that solving RL problems can be intractable as well \cite{Whitehead1991}. This work also presented that the behavior of an agent followed a random walk until it reached the goal for the first time. Koenig et al. also studied the effect of representation and knowledge on the tractability of goal-directed exploration with RL algorithms \cite{Koenig1996}. However, in our work, the primary focus is on the second stage of GDRLP, where the aim is to accelerate the convergence to optimal policy via model characeterization.

MF approaches in reinforcement learning can learn complex policies but requires many trials to converge. 
The most widely used model-free reinforcement learning might be the Q-Learning~\cite{Watkins92qlearning} which is detailed in the next section.
Another model-free learning algorithm similar to Q-Learning is SARSA \cite{Rumm1994}.
The main difference between SARSA and Q-Learning is that SARSA agent learns the action-value function by following the policy it learned, while Q-Learning agent learns the action-value function by following an exploitation policy owing to the exploration/exploitation trade-off.
On the other hand, MB approaches can generalize to new tasks and environments in fewer trials, however, an accurate model is necessary. We recently also investigated reachability heuristics and showed that computational performance for standard and accurate MDP models can be improved~\cite{debnath2017}.

Another interesting research direction focused on reducing the size of problem space in MB approaches. Boutilier et al. proposed structured reachability analysis of MDPs in order to remove variables from problem description, thereby reducing the size of MDP and eventually making it easier to solve \cite{Boutilier1998}.
It is therefore very intuitive to investigate approaches that combine the advantages of MF and MB methods \cite{Chebotar2017} \cite{Bansal2017}.
There have also been multiple previous works that combined the two paradigms. The primary objective of such methods were to speed-up the learning process for MF reinforcement learning approaches. A broad area of research including the DYNA framework ~\cite{Sutton90integratedarchitectures}~\cite{Sutton91planningby} leveraged a learned model to generate synthetic experience for MF learning. 

Along similar direction, several prior works focused on devising a model as an initialization for MF component \cite{Farshidian2014} \cite{Nagabandi2017}. One of the challenges that this leads to is the inaccuracies in the model which cause the issue of model bias. A suggested solution to overcome model bias is to directly train the
dynamics in a goal-oriented manner \cite{Donti2017} \cite{Bansal2017B}. Our work is also motivated from this approach in order to deal with model bias.

Unlike prior works on combining MB and MF reinforcement learning methods, we integrate the benefits of both approaches - MB and MF, with an aim to improve the sample efficiency during the training process. 
The primary objective in this work is to incorporate a model-based characterization using MFPT into a reinforcement learning algorithm (model-free approaches like Q-Learning or RL frameworks like DYNA), so that the characterization 
 result can be used to guide following MF exploration and learning.
Our approach differs from existing hybrid models in that, the method does not need to construct an accurate model, and a rough model with little cost is enough for capturing high-level features. 
By comparing with state-of-the-art baseline approaches, our evaluations reveal that the proposed hybrid algorithms are able to learn optimal policy in very few trials with high sample efficiency, and have significantly accelerated the practical learning process.


\definecolor{roweven}{rgb}{1,1,1}
\definecolor{rowodd}{rgb}{0.95,0.95,0.995}

\section{Preliminaries}
\label{Preliminaries}

\subsection{Model-Based Reinforcement Learning}

Model-Based reinforcement learning needs to first build a model, and then use it to derive a policy.
The underlying mechanism is Markov Decision Process (MDP), which is a tuple $M = (S,A,T,R)$, where $S = \{s_1, \cdots, s_n\}$ is a set of states and $A = \{a_1, \cdots, a_n\}$ is a set of actions. The state transition function $T : S \times A \times S \rightarrow [0,1]$ is a probability function such that $T_{a}(s_1,s_2)$ is the probability that action $a$ in state $s_1$ will lead to state $s_2$, 
and $R : S \times A \rightarrow \mathbb{R}$ is a reward function where  $R_a(s, s')$ returns the immediate reward received on taking action $a$ in state $s$ that will lead to state $s'$.
A {\em policy} is of the form $\pi = \{s_1 \rightarrow a_{1}, s_2 \rightarrow a_{2},\cdots, s_{n} \rightarrow a_{n} \}$. We denote $\pi[s]$ as the action associated to state $s$. 
If the policy of a MDP is fixed, then the MDP behaves as a Markov chain~\cite{kemeny1959finite}.

To solve an MDP, the most widely used approach should be value iteration (VI).
The VI is an iterative procedure that calculates the value (or utility in some literature) of each state $s$ based on the values of the neighbouring states that $s$ can directly transit to. 
The value $V(s)$ of state $s$ at each iteration can be calculated by the Bellman equation shown below
\begin{equation} \label{eq:mdp1}
V(s) = \max_{a\in A} \sum_{s'\in S} T_a( s, s') \Big(R_a(s, s')  + \gamma V(s') \Big),
\end{equation}
where $\gamma$ is a reward discounting parameter. The stopping criteria for the algorithm is when the values calculated on two consecutive iterations are close enough, i.e., 
$\max_{s \in S} |V(s) - V'(s)| \leq \epsilon$, 
where $\epsilon$ is an optimization tolerance/threshold value, which determines the level of convergence accuracy.

Relevant to this work, the {\em prioritized sweeping} mechanism  is an important heuristic-based approach for efficiently solving MDPs in order to further speed up the value iteration process ~\cite{Moore93prioritizedsweeping}.
This heuristic evaluates each state and obtains a score based on the state’s contribution to the convergence, and then prioritizes/sorts all states based on their scores (e.g., those states with larger difference in value between two consecutive iterations will get higher scores) ~\cite{parr1998generalized,wingate2005prioritization}. Then immediately in the next dynamic programming iteration, evaluating the value of states follow the newly prioritized order. 

Given a model, methods proposed for solving MDPs can be easily extended to the context of MB learning methods~\cite{KaelblingSurvey1996}. The model-based characterization  in our proposed approach is also built on top of this notion.


\subsection{Model-Free Reinforcement Learning}

Model-free reinforcement learning aims at learning a policy without learning a model.
The most widely used model-free reinforcement learning might be the Q-Learning~\cite{Watkins92qlearning}, which is a special algorithm of the Temporal Difference (TD) learning~\cite{Sutton1988}.
This approach is able to compare the expected utility of the available actions at a given state without requiring a model of the environment. To learn the expected utility of taking a given action $a$ in a given state $s$, it learns a action-value function $Q(s,a)$. The Q-Learning rule is
\begin{equation} \label{eq:ql}
Q(s,a) = Q(s,a) + \alpha (r + \gamma \max_{a'\in A} Q(s',a') - Q(s,a)),
\end{equation}
where, <$s,a,r,s'$> is an experience tuple, $\alpha$ is the learning rate, and $\gamma$ is the discount factor. After the action-value function is learned, the optimal policy can be constructed by greedily selecting the action with the highest action-value in each state.



\subsection{Synthesis of Model-based and Model-free}

There are several frameworks that integrate the model-based and model-free paradigms, with the most well known architecture probably being DYNA~\cite{Sutton90integratedarchitectures,Sutton91planningby}. DYNA exploits a middle ground, yielding strategies that are both more effective than model-free learning and more computationally efficient than the certainty-equivalence approach. DYNA architecture comprises of two phases. In the first phase, the agent carries out actions in the environment and performs regular reinforcement learning to learn value function and adjust the policy. It also uses the real experience to explicitly build up the transition model $T$ and/or the reward function $R$ associated with the environment. The second phase involves planning updates where simulated experiences are used to update policy and value function.


\begin{figure}[t]
  \centering
  \subfigure[]
        {\label{fig:S_Env}\includegraphics[height=1.5in]{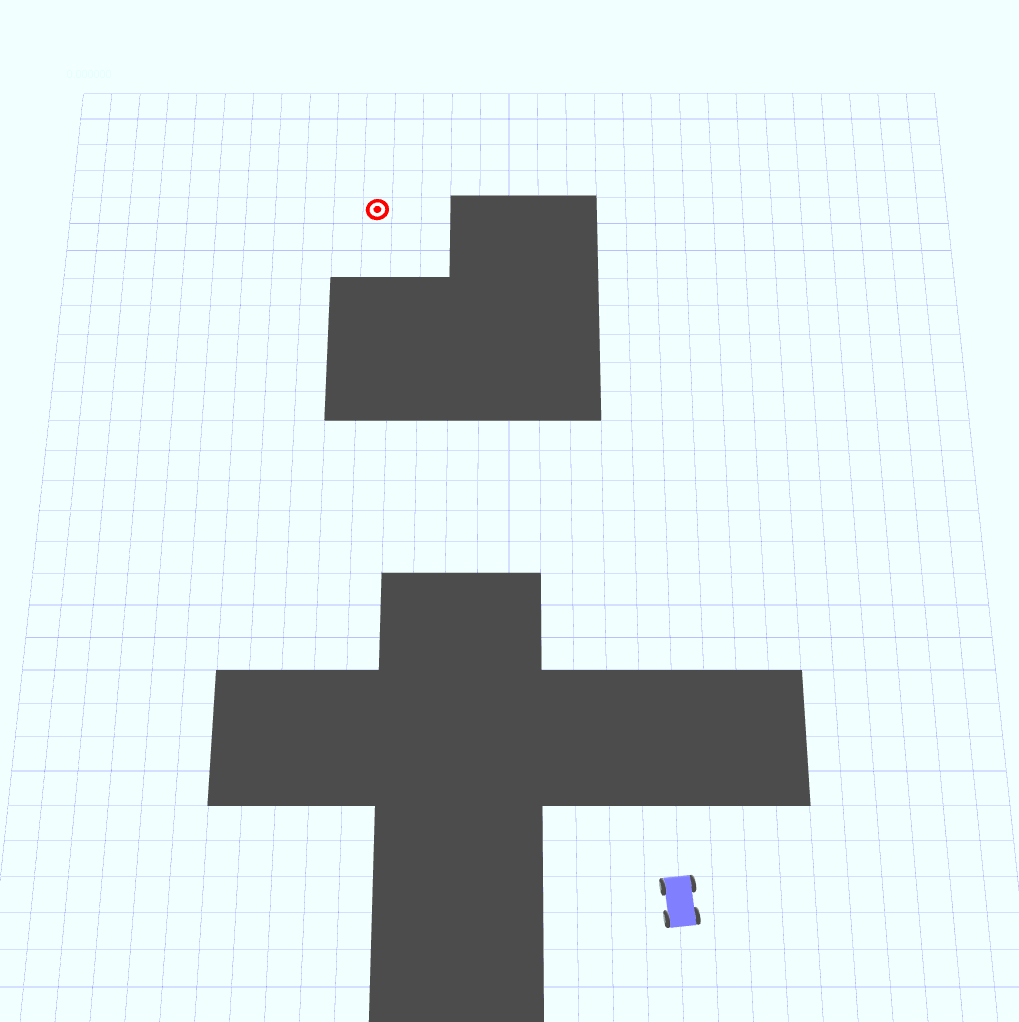}}
        \quad 
  \subfigure[]
        {\label{fig:trajectory}\includegraphics[height=1.5in]{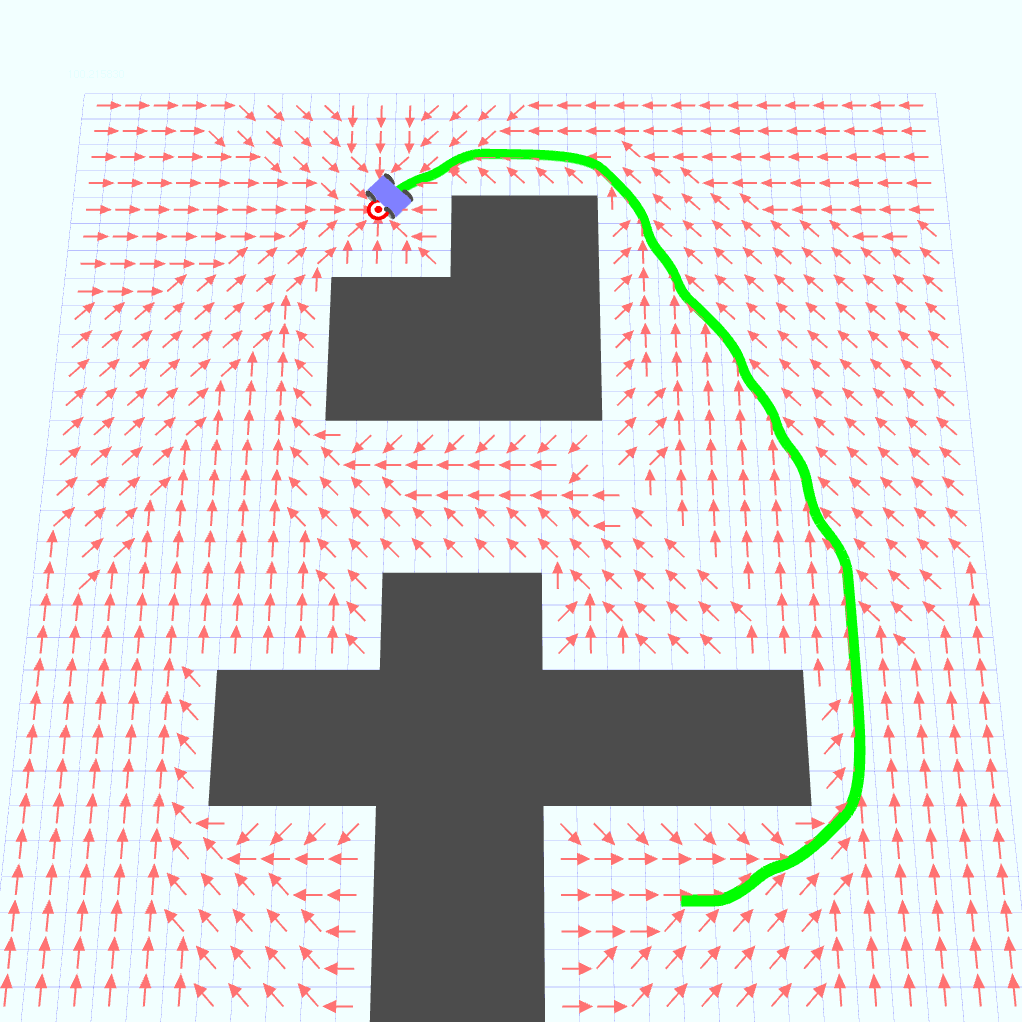}}\\
  \subfigure[]
        {\label{fig:heatmap01}\includegraphics[height=0.65in]{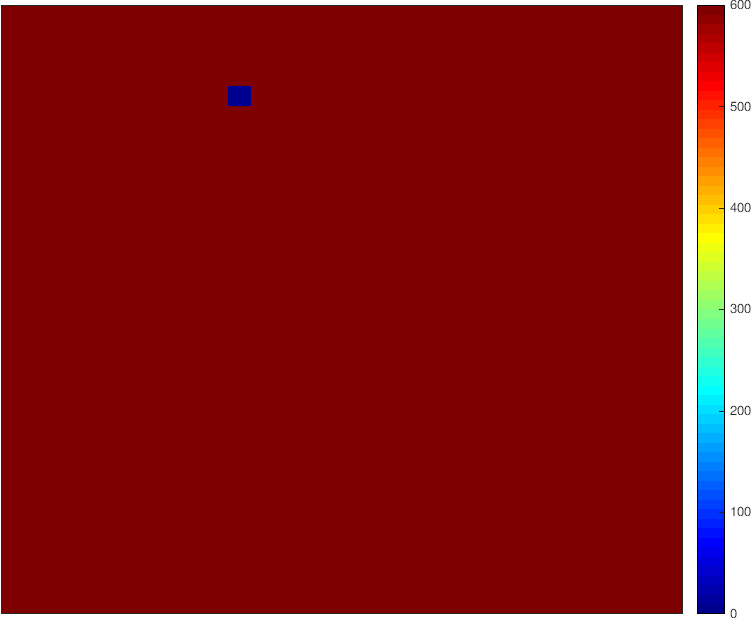}}
  \subfigure[]    
        {\label{fig:heatmap11}\includegraphics[height=0.65in]{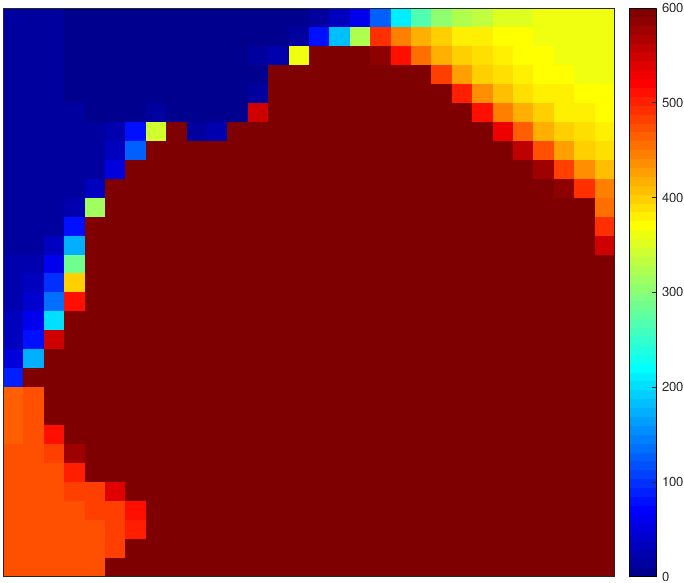}} 
  \subfigure[]
        {\label{fig:heatmap21}\includegraphics[height=0.65in]{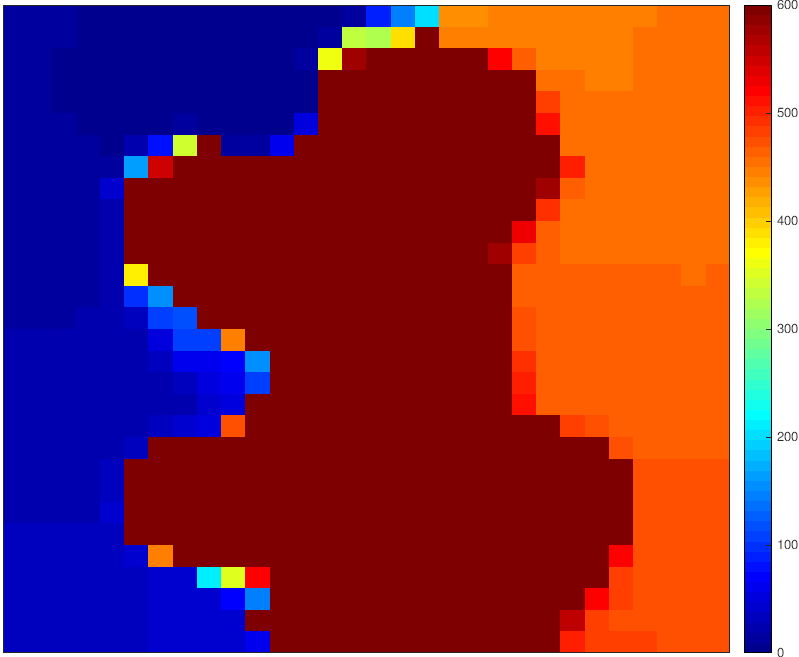}}  
  \subfigure[]
        {\label{fig:heatmap41}\includegraphics[height=0.65in]{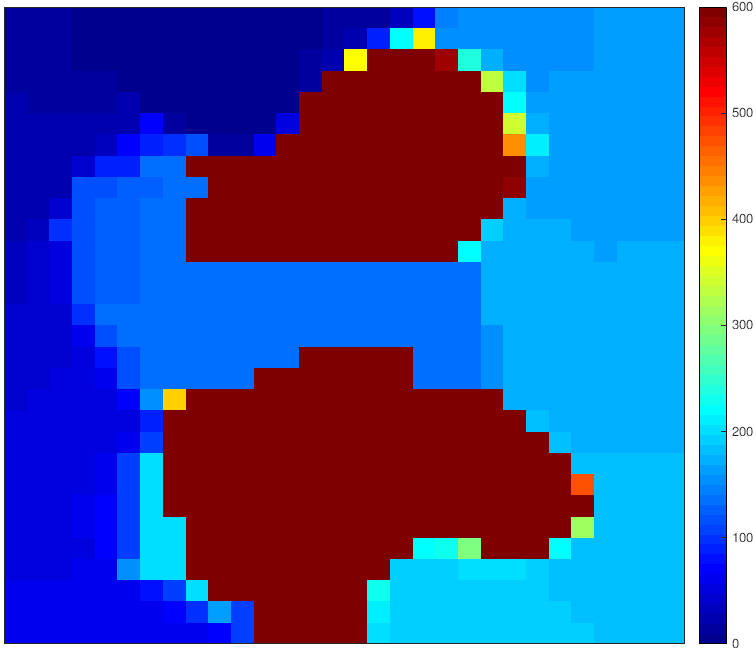}}
	\caption{\small (a) Demonstration of simulation environment, with the agent's initial state (blue) and the goal state (red). Grey blocks are obstacles; (b) Converged optimal policy (red arrows) and a trajectory completed by the agent to reach the goal.; (c)-(f) Evolution of  reachability landscapes.}
	\label{fig:heatmapApp}
\end{figure}


\subsection{Mean First Passage Times}

Before we elaborate on what we mean by model characterization in our work, we will describe a key concept called Mean First Passage Times (MFPT).

The {\em first passage time (FPT)}, $T_{ij}$, is defined as the number of state transitions involved in reaching states $s_j$ when started in state $s_i$ for the first time. The {\em mean first passage time (MFPT)}, $\mu_{ij}$ from state $s_i$ to $s_j$ is the expected number of hopping steps for reaching state $s_j$ given initially it was in state $s_i$ ~\cite{AssafSharedShanthikumar1985}. 
The MFPT analysis is built on the Markov chain, and has nothing to do with the agent's actions. 
Remember that, when a MDP is associated to a fixed policy, it then behaves as a Markov chain~\cite{kemeny1959finite}. 

Formally, let us define a Markov chain with $n$ states and transition probability matrix, $p \in {\rm I\!R}^{n,n}$. If the transition probability matrix is regular, then each MFPT, $\mu_{ij} = E(T_{ij})$, satisfies the below conditional expectation formula:
\begin{equation}\label{eq:fpt00}
E(T_{ij}) = \sum_{k} E(T_{ij} | B_k) p_{ik},
\end{equation}
where, $p_{ik}$ represents the transition probability from state $s_i$ to $s_k$, and $B_k$ is an event where the first state transition happens from state $s_i$ to $s_k$. 
From the definition of mean first passage times, we have, $E(T_{ij} | B_k) = 1 + E(T_{kj})$. So, we can rewrite Eq.~\eqref{eq:fpt00} as follows
\begin{equation}\label{eq:fpt01}
E(T_{ij}) = \sum_{k} p_{ik} + \sum_{k \neq j} E(T_{kj}) p_{ik}.
\end{equation}
Since, $\sum_{k} p_{ik}$ = 1, Eq.~\eqref{eq:fpt01} can be formulated as per the below equation:
\begin{equation}\label{eq:fpt}
\mu_{ij} = 1 + \sum_{k \neq j} p_{ik} * \mu_{kj} \quad \Rightarrow \quad \sum_{k \neq j} p_{ik} * \mu_{kj} - \mu_{ij} = -1,
\end{equation}
Solving all MFPT variables can be viewed as solving a system of linear equations 
\begin{equation}
\label{eq:fpt2}
\begin{spmatrix}{}
    p_{11} - 1 & p_{12} & .. & .. & p_{1n} \\        
    p_{21} & p_{22} - 1 & .. & .. & p_{2n} \\
    .. & .. & .. & .. & .. \\
    .. & .. & .. & .. & .. \\
    p_{n1} & p_{n2} & .. & .. & p_{nn} - 1 \\
\end{spmatrix}
\begin{spmatrix}{}
    \mu_{1j} \\        \mu_{2j} \\ .. \\ .. \\ \mu_{nj}
\end{spmatrix}
=
\begin{spmatrix}{}
    -1  \\        -1 \\ .. \\ .. \\ -1
\end{spmatrix}.
\end{equation}
The values $\mu_{1j}$, $\mu_{2j}$, $....$, $\mu_{nj}$ represents the MFPTs calculated for state transitions from states $s_1$, $s_2$, $....$, $s_n$ to $s_j$ and $\mu_{jj} = 0$. 
To solve above equation, efficient decomposition methods~\cite{Golub1996} may help to avoid a direct matrix inversion.

\section{Technical Approach}
\label{technical}

In this work, we are interested in goal-directed autonomy, where the agent is specified with a goal or terminal state to arrive.
Note, a Markov system is defined as {\em absorbing} if from every non-terminal state it is possible
to eventually enter a goal state~\cite{boutilier2000stochastic}. 
We restrict our attention to absorbing Markov systems so that the agent finally terminates at a goal.


\subsection{Reachability Characterization using Mean First Passage Times}

The notion of MFPT allows us to define the {\em reachability} of a state.
By ``reachability of a state" we mean that based on current fixed policy, how hard it is for the agent to transit from the current state to the given goal/absorbing state. 
With all MFPT values $\mu_{ij}$ obtained, we can construct a {\em reachability landscape} which is essentially a ``map" measuring the {\em degree of difficulty} of all states transiting to the goal.

Fig.~\ref{fig:heatmapApp} shows a series of landscapes represented in heatmap in our simulated environment. 
The values in the heatmap range from 0 (cold color) to 600 (warm color). 
In order to better visualize the low MFPT spectrum that we are most interested, any value greater than 600 has been clipped to 600. Fig.~\ref{fig:heatmap01}-\ref{fig:heatmap41} show the change of landscapes as the learning algorithm proceeds. 
Initially, all states except the goal state are initialized as unreachable, as shown by the high MFPT color in Fig.~\ref{fig:heatmap01}. 

We observe that the reachability landscape conveys very useful information on potential impacts of different states. 
More specifically, \textbf{a state with a better reachability (smaller MFPT value) is more likely to make a larger change during the MDP convergence procedure, leading to a bigger convergence step.}
With such observation and the new metric, we can design state prioritization schemes for Value Iteration that we will use in our proposed approaches.



\begin{algorithm}
    \caption{Mean First Passage Time based Q-Learning (MFPT-Q)}
    \label{algo:fptq}
    {\small
        Given states $S$, actions $A$, discount factor $\gamma$, learning rate $\alpha$ and goal state $s^*$,
        calculate the optimal policy $\pi$\\
                
        \While{true}{
        
            Select state $s$ at random
            
            Choose action $a$ based on $\epsilon$-greedy
            
            Execute $a$ at state $s$ and get $s'$, $r$
            
            Perform one-step tabular Q-Learning:
            $Q(s,a) = Q(s,a) + \alpha[r + \gamma \max_{a'} Q(s',a') - Q(s,a)]$
            
            Update model details: transition probability $T_{a}(s,s')$ and reward $R_{a}(s,s')$ based on count statistics
        
            \While{true}{
                $V = V'$ \\
                
                
                Calculate MFPT values $\mu_{1s^*}$, $\mu_{2s^*}$, $\cdots$, $\mu_{|S|s^*}$ by solving the linear system as shown in Eq. ~\eqref{eq:fpt2} \\
                
                List $L$ := Sorted states with increasing order of MFPT values \\
                
                \ForEach{state $s$ in $L$} {
                    Compute value update at each state $s$ given policy $\pi$:  
                    $Q(s, a) = \sum_{\forall{s'\in S}} T_a( s, s') \Big(R_a(s, s')  + \gamma V(s') \Big)$ \
                    $V'(s) = \max_{a\in A} Q(s,a)$
                }
                            
                \If{$\max_{s_i} |V(s_i) - V'(s_i)| \leq \epsilon$}{
                    break \\ 
                }
            }
          }
        }
\end{algorithm}

\subsection{Mean First Passage Time based Q-Learning (MFPT-Q)}
Classic Q-Learning converges to the optimal solution through the Q-Learning rule as shown in Eq.~\eqref{eq:ql}, which essentially learns the state-action value function $Q(s,a)$ that represents the expected utility of the available actions at a given state.

Our proposed hybrid algorithm, MFPT-Q, performs two main operations involving a model-free and a model-based update every iteration. Firstly, given an experience <$s,a,r,s'$>, it builds an approximate model by updating the transition function $T_a(s,s')$ and reward function $R_a(s,s')$. To update $T_a(s,s')$, it uses the \textit{count statistics} which basically refers to the count of transitions occurring from state $s$ to $s'$ followed by normalization between 0 and 1. 

The second step leverages the approximate model computed earlier to perform a model-based update using MFPT-VI algorithm (See lines 8-15 of Alg.~\ref{algo:fptq}). The MFPT-VI method is built on a metric using the reachability (MFPT values) since, the reachability characterization of each state reflects the potential impact/contribution of this state. Hence, such characterization provides a natural basis for state prioritization while performing a version of value-iteration for Q values during model-based update.

Note that, since the MFPT computation is relatively expensive, and the purpose of using MFPT is to characterize global and general (instead of fine) features of all states, thus it is not necessary to compute the MFPT at every iteration, but rather after every few iterations. For those iterations between two adjacent MFPT updates, the value of all states converge from a ``local refinement" perspective. 
The computational process of MFPT-Q is pseudo-coded in Alg.~\ref{algo:fptq}.



\begin{algorithm}[ht]
    \caption{Mean First Passage Time based DYNA (MFPT-DYNA)}
    \label{alg:fptd}
    {\small
        Given states $S$, actions $A$, discount factor $\gamma$, learning rate $\alpha$ and goal state $s^*$,
        calculate the optimal policy $\pi$\\
                
        \While{true}{
        
            Select state $s$ at random
            
            Choose action $a$ based on $\epsilon$-greedy
            
            Execute $a$ at state $s$ and get $s'$, $r$ (real experience)
            
            Perform one-step tabular Q-Learning:
            $Q(s,a) = Q(s,a) + \alpha[r + \gamma \max_{a'} Q(s',a') - Q(s,a)]$
            
            Update model details: transition probability $T_{a}(s,s')$ and reward $R_{a}(s,s')$ based on count statistics
            
            i $\rightarrow$ 0
            
            \While{i < N}{
                $s$ is a randomly allocated previously observed state
                
                $a$ is a random action previously carried out in s
                
                Produce simulated experience to get $s'$, $r$
                
                $Q(s,a) = Q(s,a) + \alpha[r + \gamma \max_{a'} Q(s',a') - Q(s,a)]$
                
            }
        
            \While{true}{
                $V = V'$ \\
                
                
                Calculate MFPT values $\mu_{1s^*}$, $\mu_{2s^*}$, $\cdots$, $\mu_{|S|s^*}$ by solving the linear system as shown in Eq. ~\eqref{eq:fpt2} \\
                
                List $L$ := Sorted states with increasing order of MFPT values \\
                
                \ForEach{state $s$ in $L$} {
                    Compute value update at each state $s$ given policy $\pi$:  
                    $Q(s, a) = \sum_{\forall{s'\in S}} T_a( s, s') \Big(R_a(s, s')  + \gamma V(s') \Big)$ \
                    $V'(s) = \max_{a\in A} Q(s,a)$
                }
                            
                \If{$\max_{s_i} |V(s_i) - V'(s_i)| \leq \epsilon$}{
                    break \\ 
                }
            }
          }        
    }
\end{algorithm}


\subsection{Mean First Passage Time based DYNA (MFPT-DYNA)}

Classic DYNA architecture balances between real and simulated experience to speed up the training process. As mentioned earlier, 
the agent learns a value function and updates the policy using both sets of experiences. In addition, the agent also learns a 
model of the environment using real experiences. This notion of learned model in DYNA makes it intuitive and easier to integrate our proposed 
model-based characterization.
Here we further extend this framework and propose an upgraded hybrid algorithm - Mean First Passage Time based DYNA (MFPT-DYNA). 

Remember that the classic DYNA algorithm includes two steps involving real experience and simulated experience. 
For the simulated experience, we employ the classic DYNA procedure where the simulation performs $N$ additional updates, i.e., it chooses $N$ state-action pairs at random and update state-action values according to the rule mentioned in Eq.~\eqref{eq:ql}. (See lines 9-13 of Alg.~\ref{alg:fptd}.)

Different from the standard DYNA mechanism, our MFPT-DYNA utilizes the real experience <$s,a,r,s'$> to build the approximate model by updating state-action values according to the rule mentioned in Eq.~\eqref{eq:ql}. Again, for updating $T_a(s,s')$, it increments the \textit{count statistics} for the transitions occurring from state $s$ to $s'$  followed by normalization between 0 and 1. It updates the $R_a(s,s')$ based on the reward $r$ received for taking action $a$ in state $s$.

Analogous to the one mentioned in case of MFPT-Q, we propose a model-based update using MFPT-VI algorithm. The MFPT-VI algorithm leverages the approximate model (represented by the transition and reward function) computed earlier to perform a version of the value-iteration update for $Q$ values as shown in lines 14-21 of Alg.~\ref{alg:fptd}.

One advantage of this framework is the introduction of model-based characterization by MFPT-VI. MFPT-VI very  well  assesses  the importance  of  states  based on their MFPT values  and  thus  provides  a  natural  basis  for effectively  prioritizing  states while updating state-action $Q$ values. 
This mechanism towards updating state-action $Q$ values allow the algorithm to converge with a very small number of iterations, which practically decrease the overall training time by a significant margin.
The evaluation details are presented in Section~\ref{Results}.


\subsection{Time Complexity Analysis}

Q-Learning and DYNA have a sample complexity $O(|S| \log |S|)$, where $|S|$ denotes the number of states in order to obtain a policy arbitrarily close to the optimal one with high probability~\cite{Kearns1999}~\cite{Cobo2013}.
Calculation of the MFPT needs to solve a linear system that involves matrix inversion (the matrix decomposition has a time complexity of $O(|S|^{2.3})$ if state-of-the-art algorithms are employed~\cite{Golub1996}, given that the size of matrix is the number of states $|S|$). 
Therefore, for each iteration, the worst-case time complexity for both the MFPT-Q and MFPT-DYNA algorithms is $O(|S|^{2.3})$.
Note, in practical applications, since the expensive MFPT is used for summerizing global features, this part is usually used sparsely (less frequently) which also saves much time for computation. 


\begin{figure}[t]
  \centering
  \subfigure[]
        {\label{fig:3D_Intro_1}\includegraphics[height=1.6in]{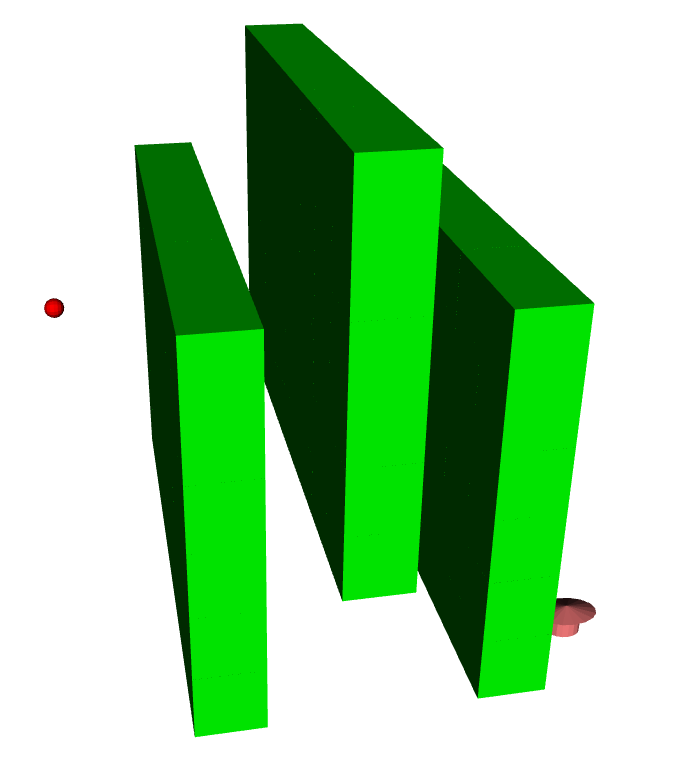}}
        \quad \quad
  \subfigure[]
        {\label{fig:3D_Intro_2}\includegraphics[height=1.6in]{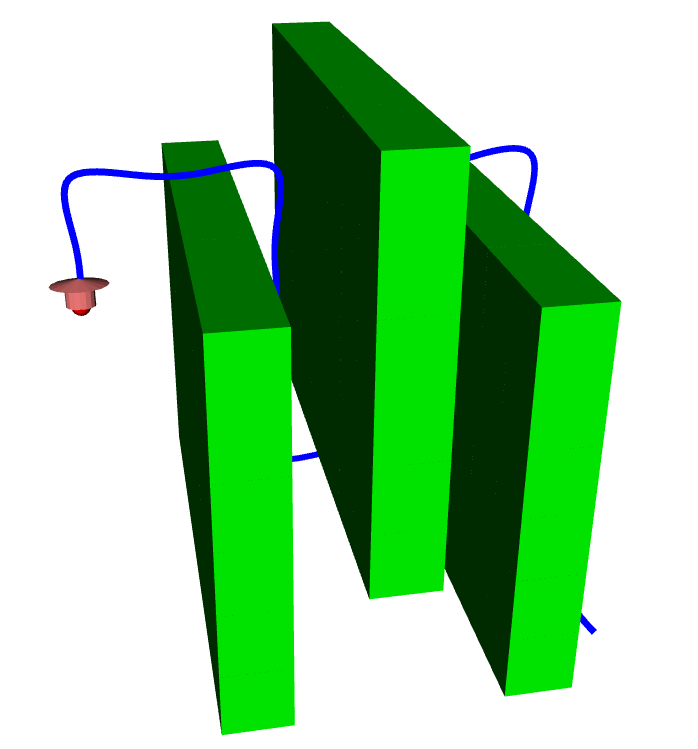}}
        \quad \quad 
	\caption{\small (a) Demonstration of simulation environment, with the agent's initial state (pink) and the goal state (red). Green blocks are obstacles; (b) A trajectory (blue) completed by the agent to reach the goal.}
	\label{fig:3DIntro}
\end{figure}

\begin{figure*}[t]
  \centering
  \subfigure[]
        {\label{fig:TS_Q}\includegraphics[height=1.4in]{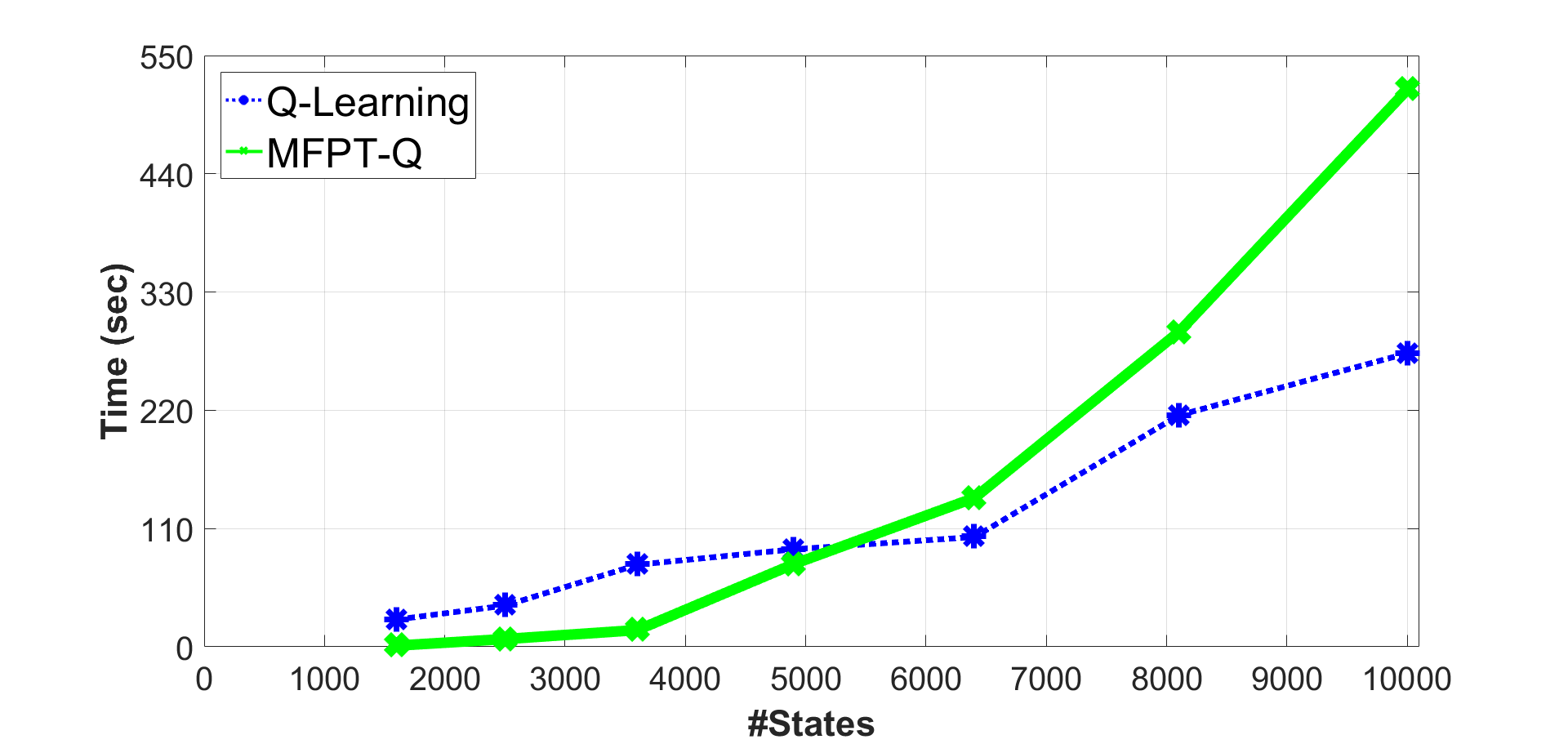}}
        \quad
  \subfigure[]    
        {\label{fig:TS_DYNA}\includegraphics[height=1.4in]{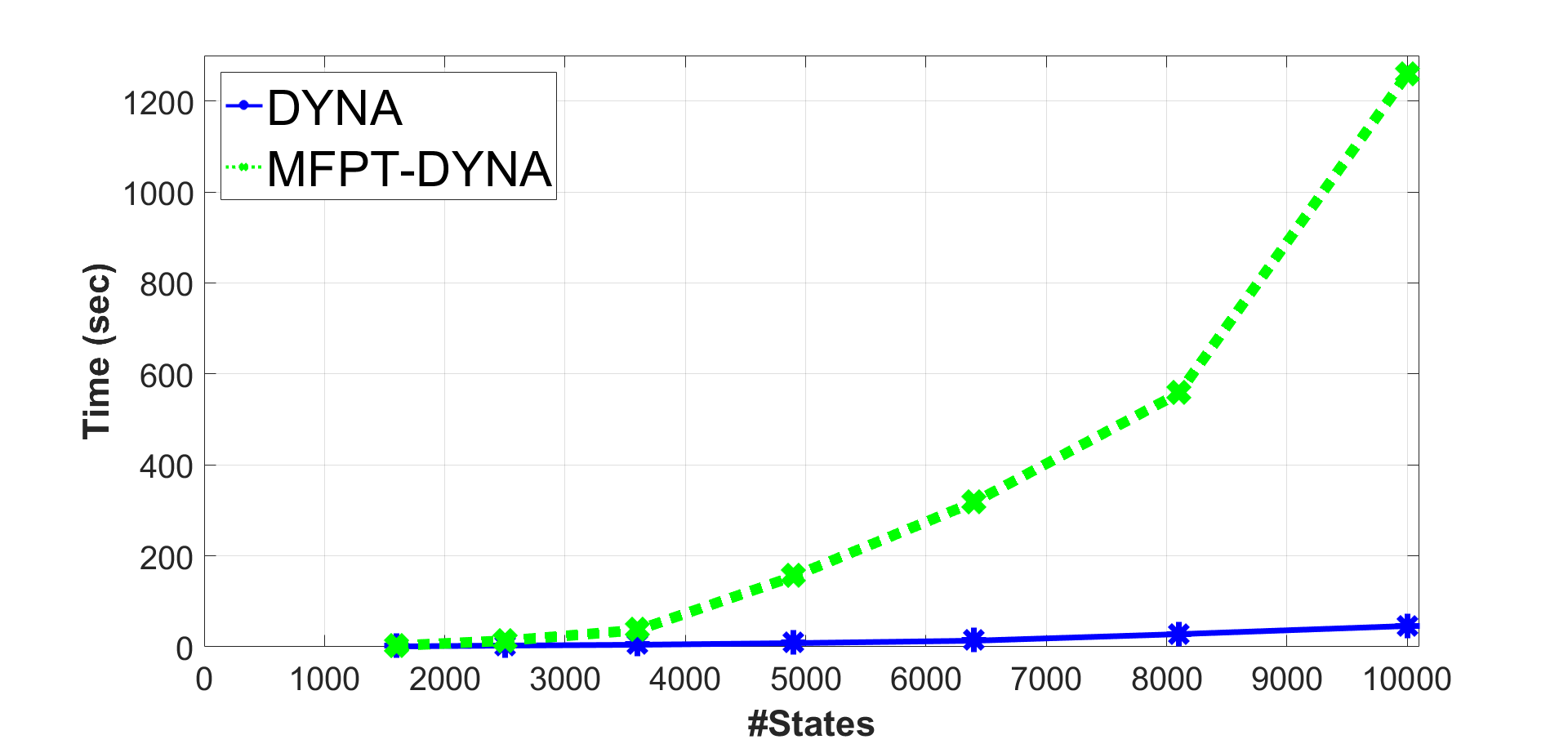}} \\
    \subfigure[]
        {\label{fig:CS_Q}\includegraphics[height=1.4in]{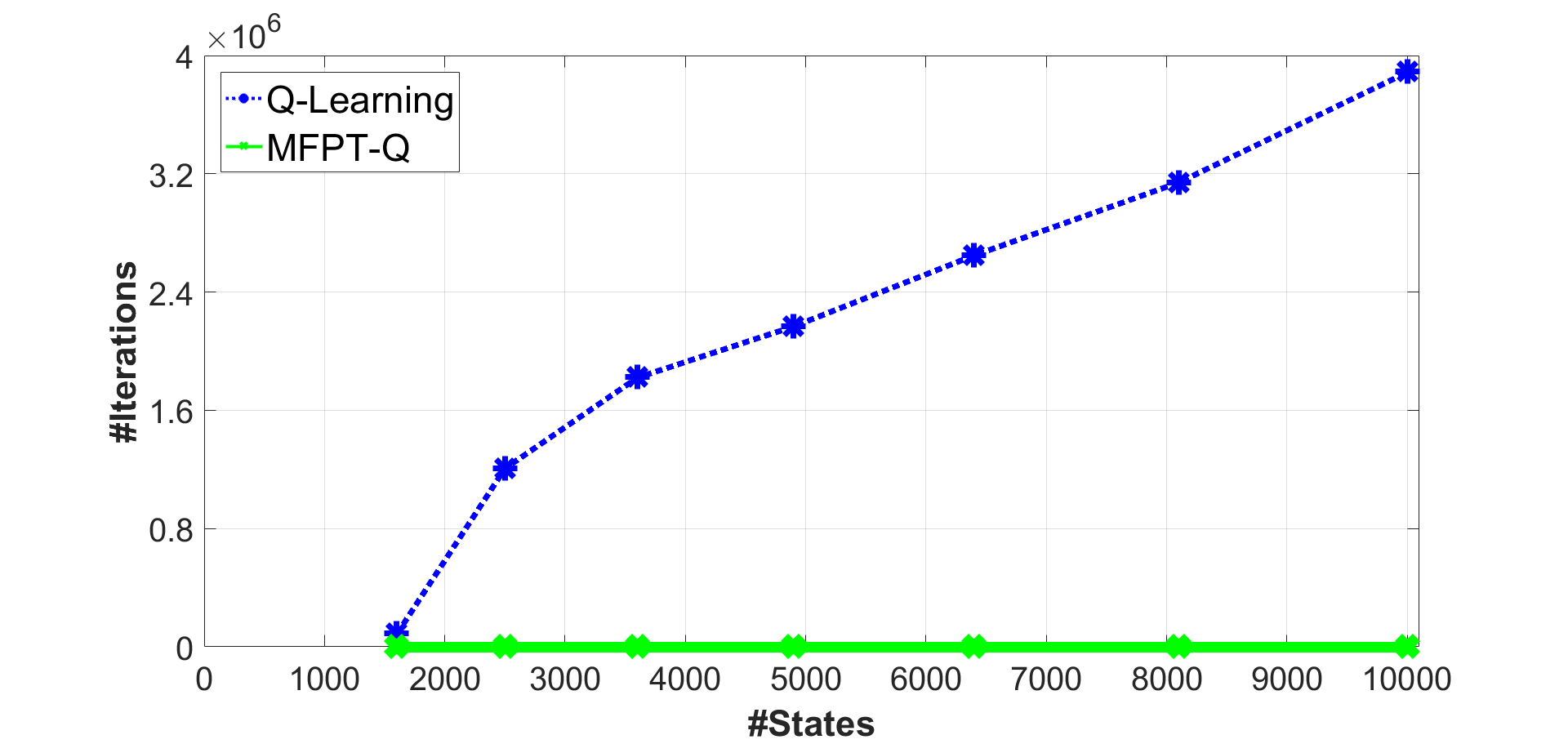}}
        \quad
    \subfigure[]    
        {\label{fig:CS_DYNA}\includegraphics[height=1.4in]{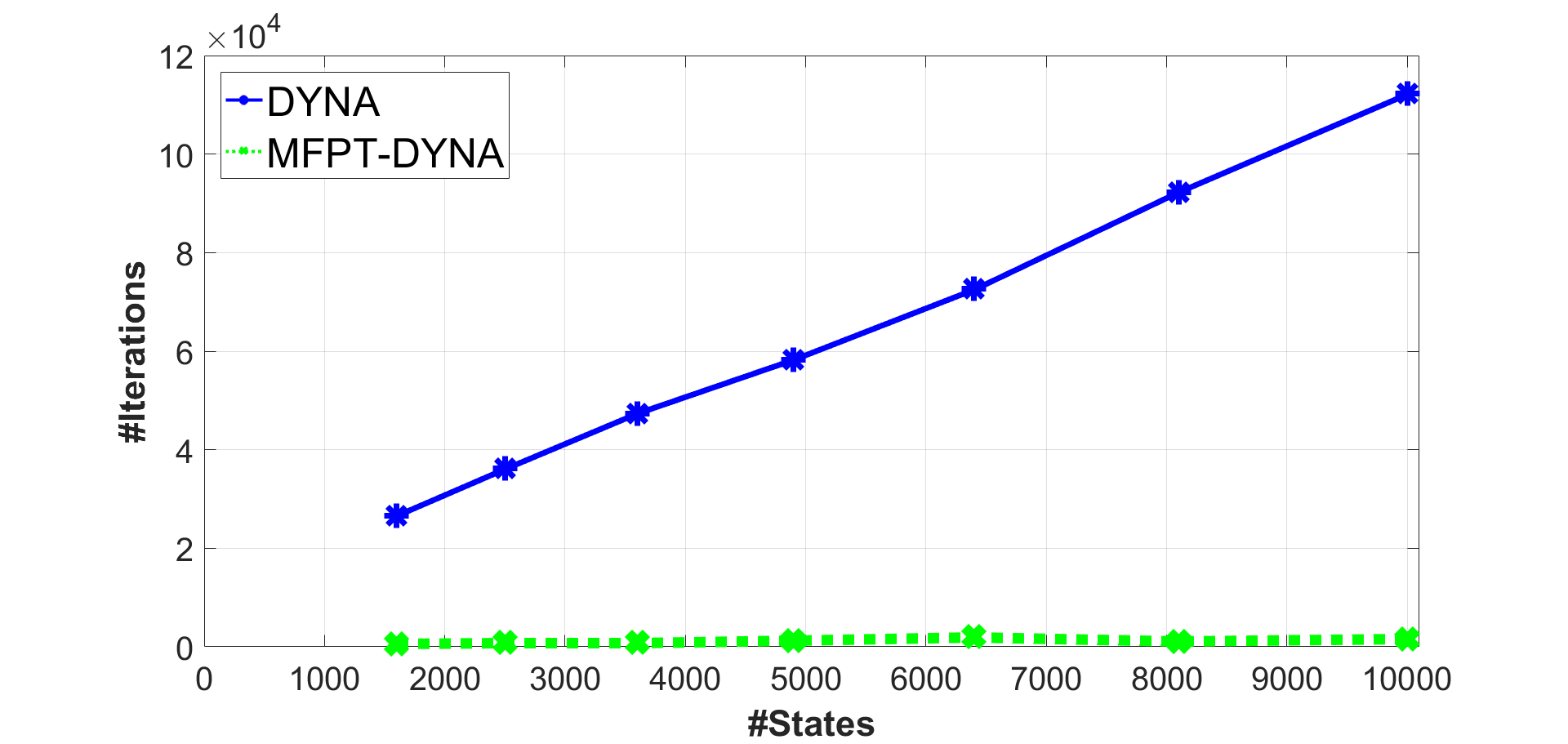}}
	\caption{\small Statistics of computational time and number of iterations required to converge between the baseline methods and our proposed algorithms, with changing numbers of states ($x$-axis) in the 2D simulation environment. (a) and (c) Variants of Q-Learning methods. (b) and (d) DYNA variants. The thick green curve in two figures is the result of MFPT-based hybrid models.
	}
	\label{fig:TimeConVsStates}
\end{figure*}

\section{Experimental Results}
\label{Results}

In this section, we compare the performance of our proposed MFPT-Q and MFPT-DYNA with their corresponding baseline methods - Q-Learning and DYNA respectively. 
More importantly, through the comparison, we wish to demonstrate that our proposed 
feature characterization mechanism can be used as a module to existing other popular frameworks (not limited to Q or DYNA) in order to further speed up their practical learning processes.


\subsection{Experimental Setting}

\subsubsection*{\textbf{Task Details}}
We validated our method through numerical evaluations with two types of simulation suites running on a Linux machine.

For the first task, we developed a simulator in C++ using OpenGL. 
To obtain the discrete MDP states, we tessellate the agent's workspace into a 2D grid map, and represent the center of each grid as a state.
In this way, the state hopping between two states represents the corresponding motion in the workspace. Each non-boundary state has a total of nine actions, i.e., a non-boundary state can move in the directions of N, NE, E, SE, S, SW, W, NW, plus an idle action returning to itself.
A demonstration is shown in Fig.~\ref{fig:heatmapApp}.
Such environmental setting allows us to better visualize the characterized reachability landscape, with a small number of states in 2D space.

For the second task, we developed a simulator in C++ using ROS and Rviz~\cite{rvizBib}. The agent's workspace is partitioned into a 3D voxel map where the center of each voxel denotes a MDP state. Each non-boundary state has a total of seven actions, i.e., a non-boundary state can move in the directions of N, E, S, W, TOP, BOTTOM plus an idle action returning to itself. 
Such a 3D environment setting is more complex compared to the 2D setting. Moreover, it can be leveraged to simulate various robotic path-planning application scenarios like quadrotor trajectory planning and demonstrate the practicality of our proposed algorithms for such tasks. A demonstration of the agent flying in the 3D simulation environment is shown in Fig.~\ref{fig:3DIntro}.

In both tasks, the objective of the agent is to reach a goal state from an initial start location in the presence of obstacles. The reward function for both setups is represented as high positive reward at the goal state and -1 for other states. All experiments were performed on a laptop computer with 8GB RAM and 2.6 GHz quad-core Intel i7 processor.

\subsubsection*{\textbf{Evaluation Methods}}

We are concerned about both computational performance and real-world training performance. Thus, we designed two evaluation metrics: 
\begin{itemize}
    \item For the first metric, we look into the {\em computational costs} of the proposed and baseline approaches where we investigate the number of iterations required to converge as well as the computational runtime used to generate the result. 
    \item For the second evaluation metric, we evaluate and compare the {\em actual time used for training and completing a task}. We do this because the robot needs to interact with the real world, and the time spent on training with the real world experience can be much more than computational time cost. Unsurprisingly, such saving also extends to other costs such as energy if the task can be done more quickly.
\end{itemize}



\begin{figure*}[t]
  \centering
    \subfigure[]
        {\label{fig:TS_Q_3D}\includegraphics[height=1.5in]{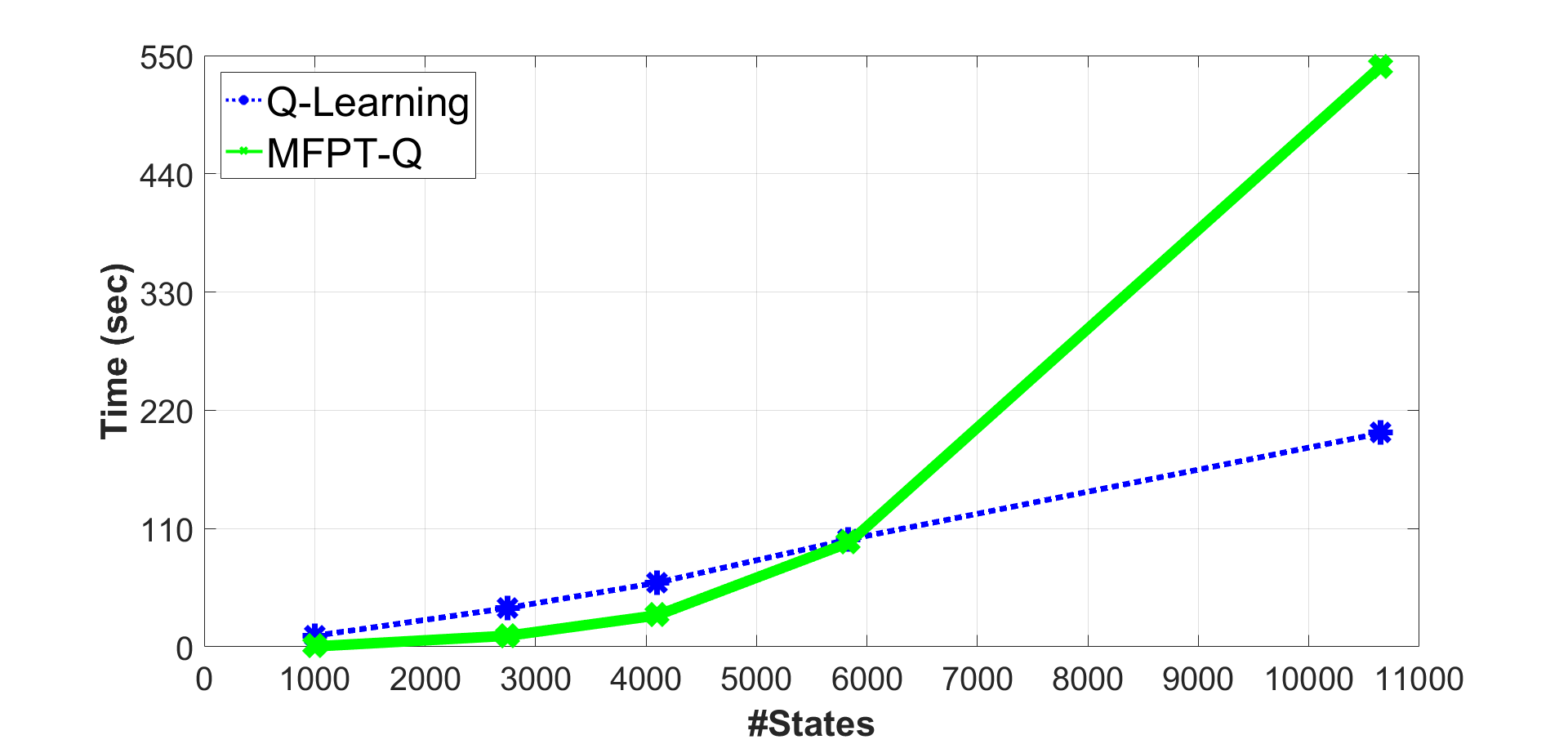}}
        \quad 
    \subfigure[]    
        {\label{fig:TS_DYNA_3D}\includegraphics[height=1.5in]{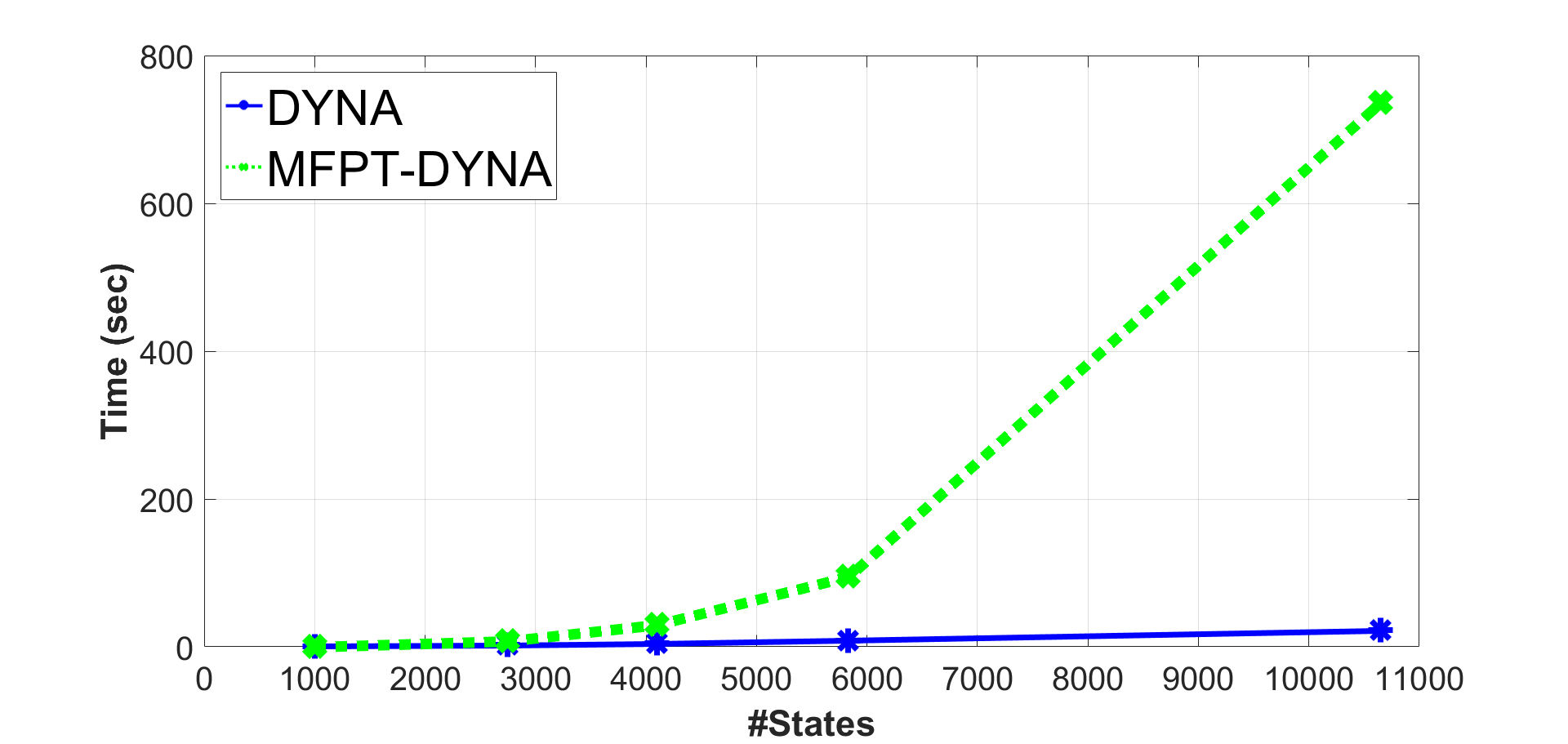}} \\
    \subfigure[]
        {\label{fig:CS_Q_3D}\includegraphics[height=1.5in]{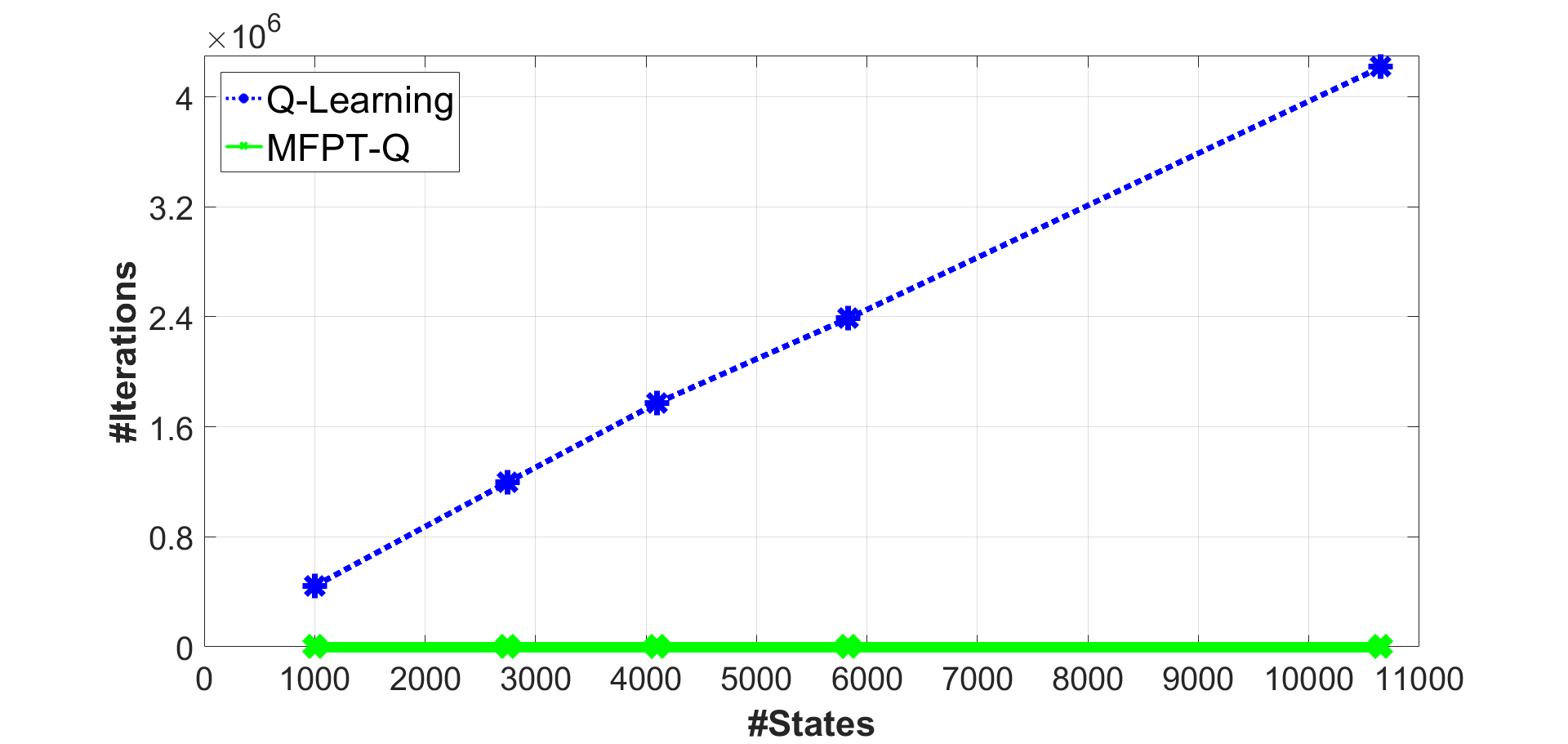}}
        \quad
    \subfigure[]    
        {\label{fig:CS_DYNA_3D}\includegraphics[height=1.5in]{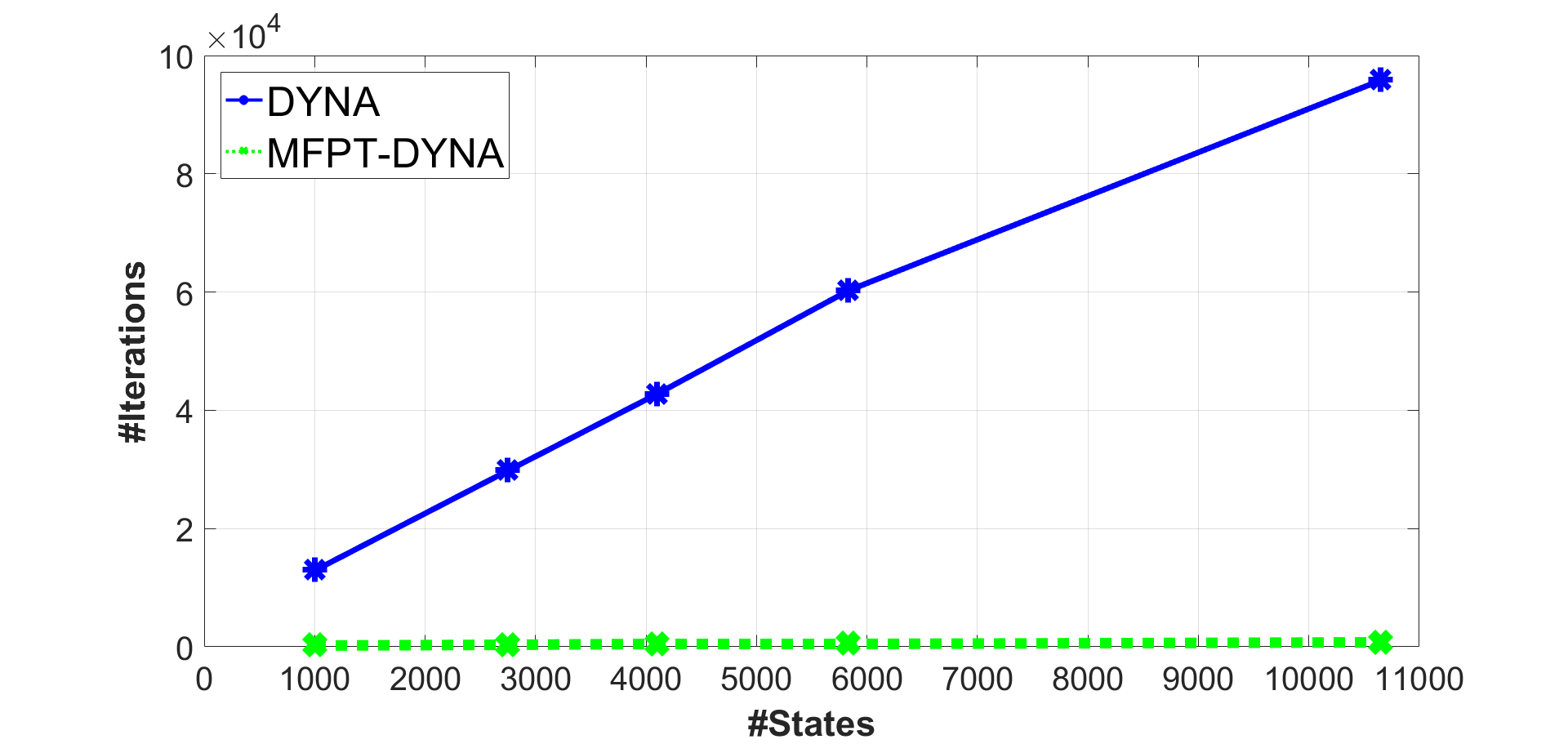}}    
	\caption{\small Comparisons of computational time  and number of iterations required to converge between the baseline methods and our proposed algorithms in the 3D simulation environments. (a) and (c) Variants of Q-Learning methods. (b) and (d) DYNA variants. 
	}
	\label{fig:TimeConVsStates_3D}
\end{figure*}

\begin{figure*}[t]
  \centering
  \subfigure[]
        {\label{fig:3D_1}\includegraphics[height=1.2in]{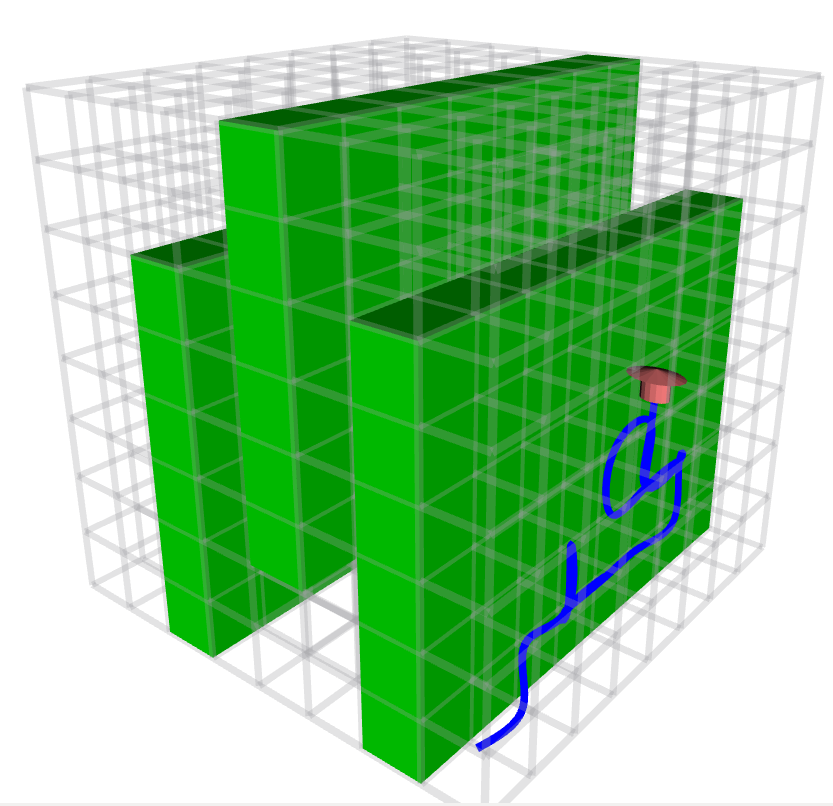}}
        \quad \quad
  \subfigure[]
        {\label{fig:3D_2}\includegraphics[height=1.2in]{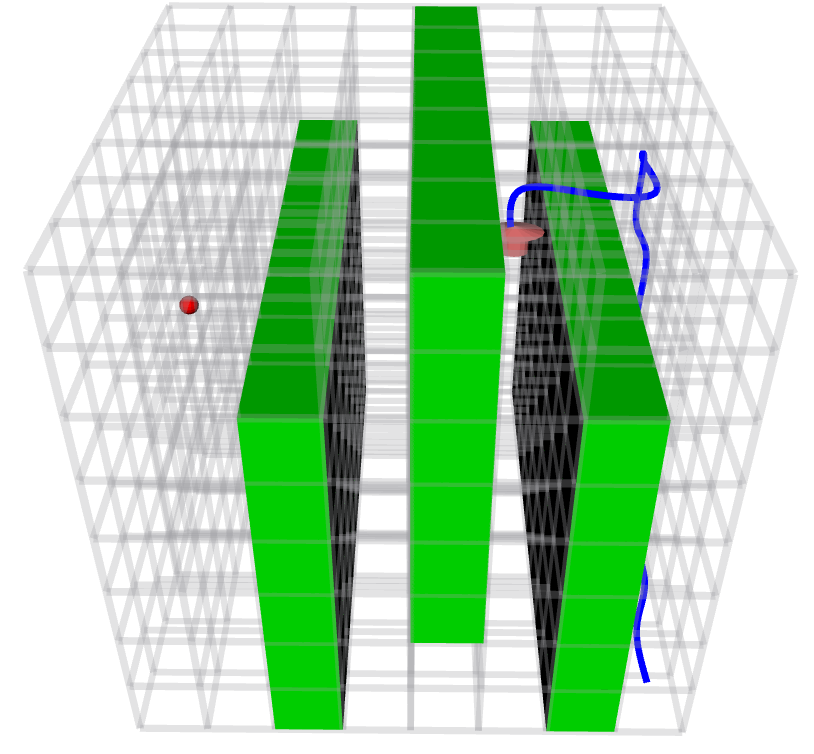}}
        \quad \quad 
  \subfigure[]
        {\label{fig:3D_3}\includegraphics[height=1.2in]{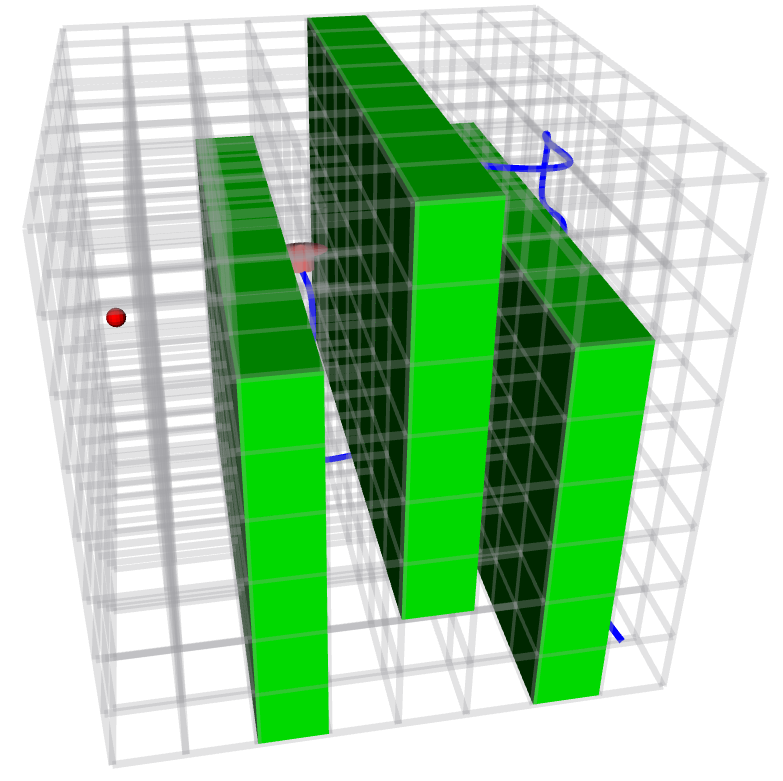}}
        \quad \quad 
  \subfigure[]
        {\label{fig:3D_4}\includegraphics[height=1.2in]{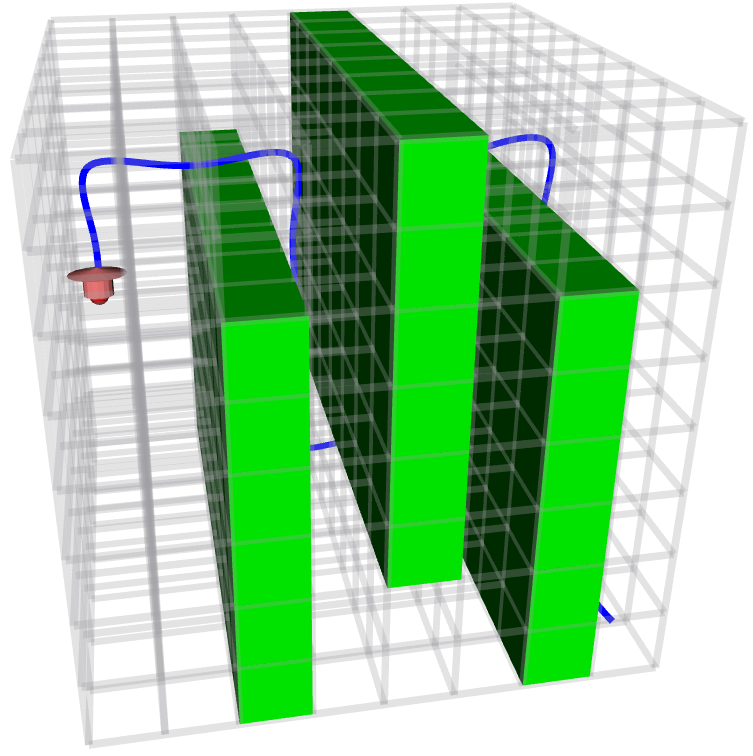}}
        \quad \quad 
	\caption{\small Agents progress during the training process via qlearning. (a) Policy learned after 500 trials; (b) Policy learned after 5000 trials; (c) Policy learned after 50000 trials; (d) Optimal policy learned by the agent. A trajectory (blue) completed by the agent to reach the goal.}
	\label{fig:3DTrajectory}
\end{figure*}

\subsection{2D Grid Setup}

In this setup, we compare our proposed algorithms with their corresponding baseline algorithms in terms of their computational runtimes as well as the numbers of iterations required to converge.

\textbf{Computational Time Cost:}
We compare the computational time taken by the algorithms as the number of states change.  
The time taken by Q-Learning and MFPT-Q algorithms to converge to the optimal solution (with the same convergence error threshold) are shown in Fig.~\ref{fig:TS_Q}. 
The results reveal that time taken by our proposed MFPT-Q to converge is faster than Q-Learning when the number of states are less than 5000. When the number of states exceed 5000, Q-Learning takes lesser time to converge than our proposed MFPT-Q. 
Fig.~\ref{fig:TS_DYNA} compares the time taken by DYNA and MFPT-DYNA.
Here, we observe that our proposed MFPT-DYNA takes more time to converge in comparison to DYNA. The reason is due to the increased time taken for MFPT calculation which dominates the time taken to converge in the execution of MFPT-DYNA.


\textbf{Number of Iterations: }
We then evaluate the number of iterations taken by the algorithms to converge to the optimal policy since it very well reflects the sample efficiency. 

Fig.~\ref{fig:CS_Q} compares the number of iterations taken by Q-Learning and MFPT-Q, respectively. The plot clearly shows that MFPT-Q takes much fewer iterations to converge in comparison to the standard Q-Learning.
Fig.~\ref{fig:CS_DYNA} compares the number of iterations taken by DYNA and MFPT-DYNA, from which we can observe that MFPT-DYNA takes much fewer itertations to converge than DYNA. 

This also implies the remarkable merit of model-based characterization via MFPT as means for faster convergence in significantly fewer iterations.



\subsection{3D Grid Setup}
To evaluate with larger number of states as well as more complex environments, we compare our proposed algorithms: MFPT-Q and MFPT-DYNA in the 3D simulation environment. 

\textbf{Computational Time Cost:}  
Fig.~\ref{fig:TS_Q_3D} presents the computational time taken by Q-Learning and MFPT-Q algorithms to converge to the optimal solution. We can see that the computational time cost trends are very similar to that of the 2D case, where for a large number of states, the MFPT variants take longer time than the baseline methods.
We attribute this to the fact that as the number of states increase, the time taken for MFPT calculation also increase which dominates the computational time cost in the MFPT variants.

\textbf{Number of Iterations: }
Similarly, we also compare the number of iterations taken by Q-Learning and MFPT-Q, respectively. As shown in Fig.~\ref{fig:CS_Q_3D}, our proposed MFPT variant converges in fewer trials compared to Q-Learning.
Next, we compare the number of iterations taken by DYNA and MFPT-DYNA. Again, the advantage of our proposed hybrid RL approach that  introduces  a model-based characterization into DYNA, is clearly visible in Fig.~\ref{fig:CS_DYNA_3D}, as the results show that the MFPT-DYNA requires much smaller number of iterations to converge compared to DYNA.



\subsection{Training Runtime Performance}

As previously discussed, the objective of RL for an agent is to learn an optimal policy in a given environment in order to reach a goal state from a given starting location. Here we present our second evaluation metric that considers the total time involved during the agent's training process in the simulation environment. 

Fig.~\ref{fig:3DTrajectory} shows the progress of an agent in the 3D  environment using Q-Learning. During the initial stages of the learning process, the agent could hardly overcome the first obstacle as shown in Fig.~\ref{fig:3D_1}. After 5000 trials, the agent could overcome the first obstacle, however was unable to overcome the next one. At the end of the training process, the agent learned the optimal policy using which it could overcome all obstacles and reach the goal as shown in Fig.~\ref{fig:3D_4}.

\begin{figure}[t]
  \centering
  \subfigure[]
        {\label{fig:Q_3D}\includegraphics[height=1.2in]{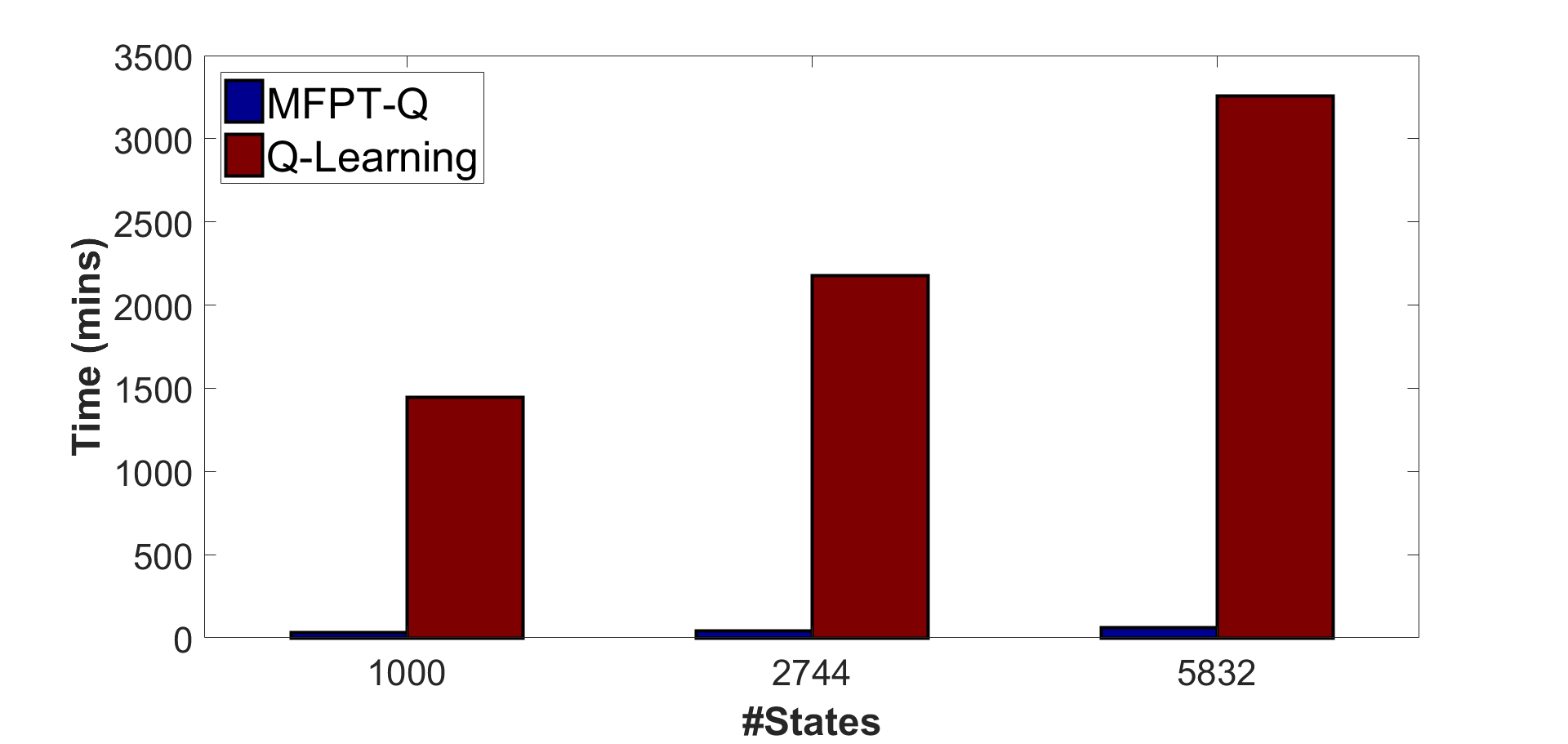}}
        \quad
  \subfigure[]    
        {\label{fig:DYNA_3D}\includegraphics[height=1.2in]{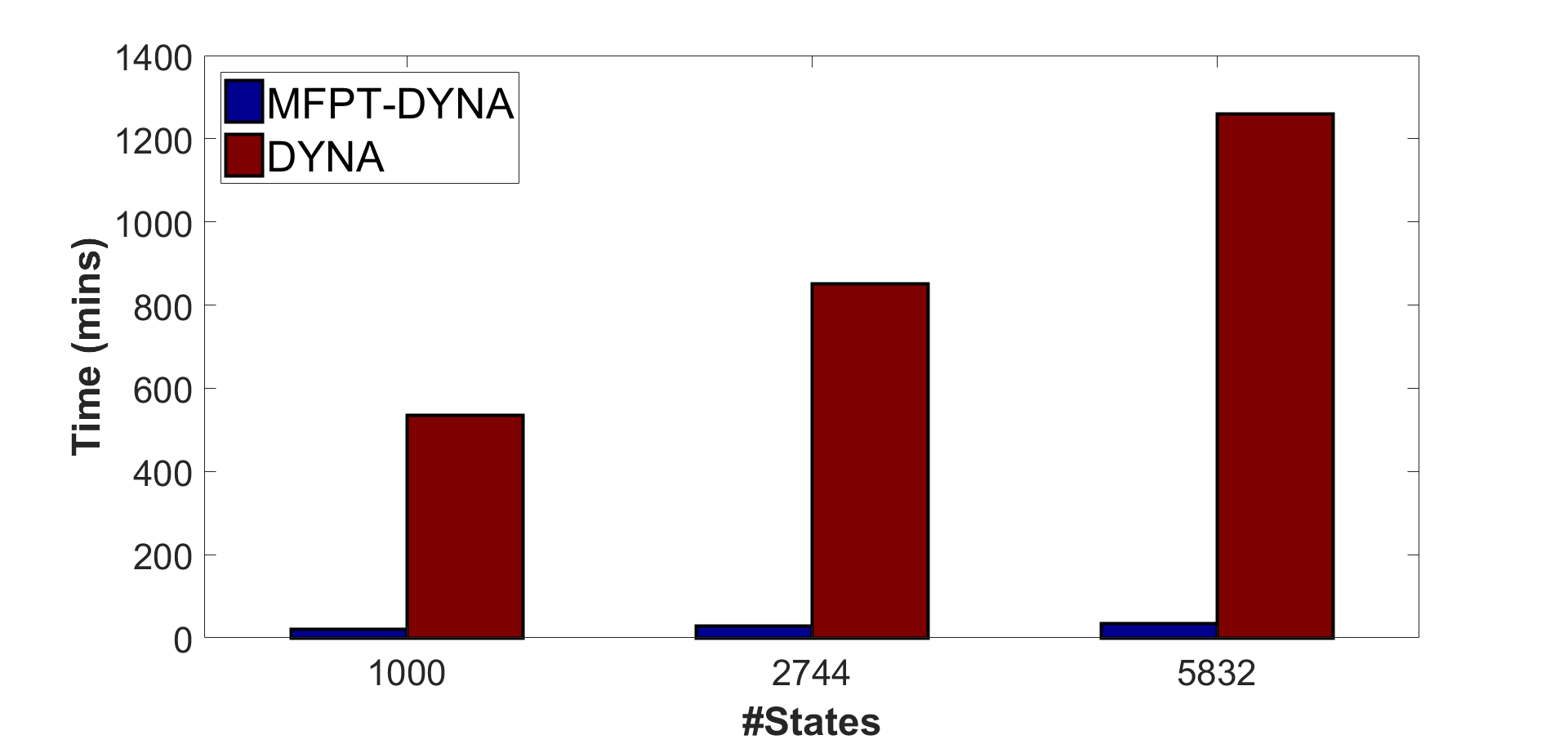}}	
	\caption{\small Time required for agent's training in the 3D environment. 
	(a) Variants of Q-Learning methods. (b) DYNA variants.
	The dark blue bar in two figures is the result of MFPT-based hybrid models.
	}
	\label{fig:Int_3D}
	\vspace{-5pt}
\end{figure}

We trained an agent in the 3D environment under varying voxel sizes. Fig.~\ref{fig:Q_3D} shows that the agent takes much less overall time to learn the optimal policy when MFPT-Q was employed in comparison to classical Q-Learning. Similarly, the agent takes much less time to learn and complete the task using MFPT-DYNA in comparison to DYNA algorithm as shown in Fig.~\ref{fig:DYNA_3D}.

Since, practically the training and learning efforts are much more expensive and important than the computational time cost, thus,  these results reestablish the benefits of our hybrid algorithms towards improving sample efficiency in goal-directed reinforcement learning.
Such faster convergence and lesser training time is owing to the underlying mechanism of model-based characterization via MFPT introduced to the existing reinforcement learning schemes. 

\section{Conclusions}
\label{conclusion}


In this paper, we propose a hybrid approach where we introduced a new model-based characterization that can be extended to reinforcement learning techniques in order to improve its sample efficiency. 
We achieved this by synthesizing the advantages of both model-free and model-based reinforcement learning paradigms. 
The proposed hybrid framework can further accelerate reinforcement learning approaches, 
via an integration of the MFPT feature characterization mechanism. 
The experimental results show the  remarkable  merit  of  model-based  characterization in our hybrid algorithms which learn much faster with fewer samples in comparison to their state-of-the-art reinforcement learning counterparts.


{
\bibliographystyle{unsrt}
\bibliography{reference}
}

\end{document}